\documentclass[lettersize,journal]{IEEEtran}
\usepackage{amsmath,amsfonts}
\usepackage{algorithmic}
\usepackage{algorithm}
\usepackage{array}
\usepackage[caption=false,font=normalsize,labelfont=sf,textfont=sf]{subfig}
\usepackage{textcomp}
\usepackage{stfloats}
\usepackage{url}
\usepackage{verbatim}
\usepackage{graphicx}
\usepackage{cite}
\usepackage{adjustbox}
\usepackage{booktabs}
\usepackage{multirow}
\usepackage{threeparttable}

\begin{document}

\title{Confident RAG: Enhancing the Performance of LLMs for Mathematics Question Answering through Multi-Embedding and Confidence Scoring}

\author{Shiting Chen, Zijian Zhao and Jinsong Chen

\thanks{This work was supported by General Research Fund by the Hong Kong Research Grants Council under Grant 17611725. \emph{(Corresponding author: Shiting Chen and Jinsong Chen.)}}
\thanks{Shiting Chen and Jinsong Chen are with the Faculty of Education, The University of Hong Kong, Hong Kong, China (e-mail: u3011355@connect.hku.hk; jinsong.chen@live.com)}
\thanks{Zijian Zhao is with the Department of Civil and Environmental Engineering, The Hong Kong University of Science and Technology, Hong Kong, China (e-mail: zzhaock@connect.ust.hk) }
}




\maketitle

\begin{abstract}
Large Language Models (LLMs) hold significant promise for mathematics education, yet they often struggle with complex mathematical reasoning. While Retrieval-Augmented Generation (RAG) mitigates these issues by grounding LLMs in external knowledge, its effectiveness remains unstable, heavily dependent on the choice of a single embedding model. Moving beyond static RAG workflows, we draw on agentic workflow patterns, a paradigm that introduces structured task decomposition and collaboration to enhance system performance. We propose and examine two novel approaches that combine the benefits of multiple embedding models. While our Mixture-Embedding RAG approach (fusing retrieved documents) shows limited gains, our Confident RAG method (generating multiple answers and selecting the one with the highest confidence score) demonstrates significant improvement. Experimental results show that Confident RAG achieved average accuracy improvements of approximately 10\% over vanilla LLMs and 5\% over vanilla RAG. The consistent results across different LLMs and embedding models indicate that Confident RAG is an efficient plug-and-play solution for trustworthy mathematical AI assistants. Finally, we discuss how this work lays the groundwork for deploying Agentic RAG systems in educational settings, where autonomous planning and iterative refinement can be built upon our robust retrieval foundation.
\end{abstract}

\begin{IEEEkeywords}
Large Language Model (LLM), Retrieval-Augmented Generation (RAG), Agentic Workflow Patterns, Mathematics Education, AI in Education.
\end{IEEEkeywords}

\section{Introduction}
\IEEEPARstart{L}{arge} language models (LLMs) have demonstrated remarkable capabilities across various domains \cite{lyu2025llm,zhao2025automatic,gao2024fine}, showing particular promise for educational applications. However, their tendency to hallucinate \cite{henkel2024retrieval} remains a significant barrier to reliable use in learning environments, especially in mathematics education where accuracy is crucial \cite{frieder2023mathematical}. In mathematical reasoning, LLMs struggle with precision, often failing to process complex symbols and formulas correctly \cite{lewkowycz2022solving}, which risks misleading students when these models are deployed as educational tools. 

Retrieval-Augmented Generation (RAG) has emerged as a powerful, plug-and-play solution to mitigate hallucination by grounding LLMs in external knowledge, which has been widely employed in constructing foundation models \cite{chen2024overview} and practical agents \cite{arslan2024sustainable}. Unlike fine-tuning, RAG enhances model performance without extensive retraining, making it both efficient and cost-effective. The main paradigm of RAG involves first calculating the similarities between a question and chunks in an external knowledge corpus, followed by incorporating the top $K$ relevant chunks into the prompt to guide the LLMs \cite{lewis2020retrieval}.

Despite the advantages of RAG, selecting the appropriate embedding models remains a crucial concern, as the quality of retrieved references directly influences the generation results of the LLM \cite{tu2025rbft}. Due to variations in training data and architecture, different embedding models capture semantic similarity in divergent ways. A model that excels on one type of mathematical problem may underperform on another, making the selection of a single optimal embedding model a difficult and uncertain choice. Consequently, this fundamental limitation makes vanilla RAG unreliable for comprehensive mathematics education applications.

Recent advancements in agentic workflow patterns offer new perspectives on enhancing RAG systems through structured task decomposition and collaborative processes \cite{singh2025agentic}. These patterns provide systematic methodologies to manage complex reasoning tasks, moving beyond static retrieval pipelines toward more adaptive and robust frameworks. 

Inspired by these developments, we address the embedding selection challenge by proposing two methods that combine the benefits of multiple embedding models for improving RAG in mathematics education. The first method is named Mixture-Embedding RAG, which sorts the retrieved materials from multiple embedding models based on normalized similarity and selects the top $K$ materials as final references. The second method is named Confident RAG, where we first utilize vanilla RAG to generate answers multiple times, each time employing a different embedding model and recording the associated confidence metrics, and then select the answer with the highest confidence level as the final response. By validating our approach using multiple LLMs and embedding models, we illustrate the superior performance and generalization of Confident RAG, providing a reliable foundation for AI-powered mathematics education. The main contributions of this paper can be summarized as follows:

\begin{itemize}
    \item We first point out that in RAG, different embedding models operate within their own prior domains, which is a critical barrier to deploying LLMs in mathematics education. To leverage the strengths of various embedding models, we propose and test two novel RAG methods: Mixture-Embedding RAG and Confident RAG. These methods effectively utilize the retrieved results from different embedding models to their fullest extent.
    \item While Mixture-Embedding RAG performs similarly to vanilla RAG, the Confident RAG method exhibits superior performance compared to both the vanilla LLM and vanilla RAG, with average improvements of 9.9\% and 4.9\%, respectively, when using the best confidence metric. Additionally, we discuss the optimal number of embedding models for the Confident RAG method based on the results.
    \item Our results reveal two outstanding confidence metrics: self-certainty and Distributional Perplexity (DP), both showing average improvements of approximately 10\% compared to the vanilla LLM. Specifically, among the LLMs examined, self-certainty achieves a maximum increase of 10.4\%, while DP demonstrates a maximum increase of 12.4\% compared to the vanilla LLM. The reasons behind the better performance of these two metrics are discussed based on their formulas.
\end{itemize}

\section{Preliminary and Related Work}

\subsection{Retrieval Augmented Generation (RAG)} \label{sec:RAG}

When constructing domain-specific foundation models or private agents, modifying a trained LLM is often necessary. However, the computational and resource costs associated with full fine-tuning render it impractical for a large number of users. Even though techniques such as Low-Rank Adaptation (LoRA) and prefix-tuning provide lighter training paradigms, RAG offers its own advantages in terms of plug-and-play capabilities. Furthermore, RAG allows external knowledge to remain independent of LLMs, facilitating the replacement of databases with other domains or more up-to-date versions.

Currently, a common paradigm of vanilla RAG is illustrated in Fig. \ref{fig:main}. Given a question $q$, an external corpus $C$ divided into chunks as $C = [c_{1}, c_{2}, \ldots, c_{m}]$, and an embedding model $\text{g}()$, we first generate the embeddings for the question and the corpus, respectively, as follows:
\begin{equation}
\begin{aligned}
e_q & = \text{g}(q) \ , \\
e_{c_j} & = \text{g}(c_j) \ ,
\end{aligned}
\label{eq:embedding}
\end{equation}
where $e_q, e_{c_j} \in \mathbb{R}^{d_g}$ are the embedding results for question $q$ and chunk $c_j$, and $d_g$ is the output dimension of the model $\text{g}()$. 

Based on the understanding that two sentences with higher similarity yield more similar embeddings, we calculate the similarity between the question and each chunk using the cosine similarity metric, defined as:
\begin{equation}
\begin{aligned}
s_{q,c_j} = \frac{e_q \cdot e_{c_j}}{\|e_q\| \|e_{c_j}\|} \ ,
\end{aligned}
\label{eq:cos-sim}
\end{equation}
where $s_{q,c_j}$ represents the cosine similarity between question $q$ and chunk $c_j$, $\cdot$ denotes the dot product, and $\|\cdot\|$ denotes the Euclidean norm. Finally, we select $k$ chunks with the highest similarity and incorporate them into the prompt, expressed as:
\begin{equation}
\begin{aligned}
\text{h}(t,q,c_j|j \in \mathcal{K}) \ ,
\end{aligned}
\label{eq:template}
\end{equation}
where $\text{h}()$ is a montage function that combines the question and the retrieved chunks using a specified template $t$, and $\mathcal{K}$ represents the index set of the selected $k$ chunks. In this way, we construct a prompt that includes external knowledge from the corpus $C$ that is most similar to question $q$. 

Previous studies have validated a wide range of modifications to the RAG pipeline, demonstrating effectiveness in open-domain question answering by improving various components such as adaptive retrieval strategies \cite{tang2024mba, guan2024external}, dynamic document selection \cite{kim2024re, hei2024dr}, joint retriever-generator optimization \cite{siriwardhana2021fine}, reinforcement learning \cite{huang2025rag}, and hybrid retrieval-reranking frameworks \cite{sawarkar2024blended, zhang2025levelrag}. 

Since embedding models can produce different similarity calculations, many studies have innovated in retriever embedding models to improve the performance of the RAG pipeline. For example, Invar-RAG \cite{liu2024invar} constructs an LLM-based retriever with LoRA-based representation learning and an invariance loss, directly addressing retriever embedding variance and locality, and demonstrates improved retrieval and response accuracy. Blended RAG \cite{sawarkar2024blended} combines dense semantic embeddings with sparse retrievers, empirically showing that this hybrid embedding space achieves higher QA accuracy, especially as corpus size scales. W-RAG \cite{nian2024w} further validates that enhancing retriever embeddings using weak signals from the downstream QA task offers gains comparable to methods requiring full human annotations. 

However, most studies have focused on the innovation of a single embedding model. This presents a critical limitation for mathematical reasoning, where effective retrieval requires a nuanced understanding of logical structures, symbols, and problem-solving strategies. Since different embedding models capture these mathematical semantics in varied and non-uniform ways, selecting a single optimal model is inherently uncertain. Our approach will leverage the strengths of various embedding models to provide the most relevant information for LLMs.


\subsection{Agentic Architectures and Workflow Patterns}
The limitations of static RAG workflows have catalyzed a paradigm shift towards embedding agentic intelligence into the retrieval-generation pipeline. This new paradigm, broadly categorized as Agentic RAG \cite{ravuru2024agentic}, moves beyond predefined sequences to systems capable of dynamic retrieval strategies, contextual understanding, and iterative refinement \cite{huang2023towards}. 

Agentic RAG is embodied by complex architectures that integrate one or more autonomous AI agents. These architectures are built upon the foundational pillars of AI agents, which include planning for decomposing complex tasks, tool use for accessing external data and APIs, reflection for iterative self-critique and refinement, and multi-agent collaboration for specialized task handling \cite{singh2025agentic}. Regarding current agentic RAG system, single-agent agentic RAG such as Self-RAG \cite{asai2024self}, Iter-RetGen \cite{shao2023enhancing}, and MetaRAG \cite{zhou2024metacognitive} embed decisions about when to query external knowledge, how to structure intermediate reasoning, and when to self-critique directly into a single LLM, enabling adaptive retrieval and reflection within one policy. Plan- and graph-centric frameworks like PAR-RAG \cite{zhang2025credible} and GraphRAG-R1 \cite{yu2025graphrag} further externalize this agency into explicit multi-step plans or reasoning graphs that guide tool calls and evidence aggregation for multi-hop questions. In contrast, multi-agent systems such as MA-RAG \cite{nguyen2025ma}, MAO-ARAG \cite{chen2025mao} and SIRAG \cite{wang2025sirag} decompose the pipeline into specialized planner, retriever, evidence-selector, and answerer agents that communicate via shared memories or chain-of-thought messages, allowing more modular and interpretable control over retrieval and generation.

Agentic workflow patterns are used to support agents to optimize performance, accuracy, and efficiency under varying task complexity and resource budgets, containing prompt chaining, routing, parallelization, orchestrator-workers, and evaluator-optimizer \cite{Anthropic2024Agents}. Prompt chaining decomposes problems into sequential subproblems, while carefully managing error propagation across steps \cite{barrak2025traceability}. Routing directs inputs or subtasks to specialized models or agents to achieve better accuracy and cost trade-offs than any single static pipeline \cite{zhang2025multi}. Parallelization stabilizes processing by running independent workflows concurrently, yielding more reliable solutions on complex tasks \cite{zhang2025optimizing}. Orchestrator–worker architectures use a controller LLM to dynamically decompose tasks, delegate them to specialized executor LLMs, and coordinate their interaction, improving reliability and adaptability compared to flat single-agent setups \cite{shi2025aime}. Finally, evaluator-optimizer iteratively critique and refine intermediate outputs, enabling smaller models to achieve larger performance at lower cost \cite{wang2025scoreflow}.

\subsection{Confidence of LLMs} \label{sec:confidence}
The confidence level of the LLM output can indicate the response quality to some extent. Several approaches have been employed to measure model confidence. Some of them depend on model prompting, external similarity or model-expressed judgments. These include linguistic confidence measures \cite{shrivastava2023llamas, xiong2023can}, where the model is prompted to state its confidence explicitly as part of the output, as well as LLM-as-judge or self-evaluation strategies that involve prompting the model to rate or compare its own answers in scalar or relative terms \cite{shrivastava2025language, ren2023self, hager2025uncertainty}. However, these methods are often poorly calibrated, not directly related to the true predictive uncertainty of the model, and can be biased or inconsistent due to their reliance on language generation rather than statistical likelihood. 

In contrast, probability-based metrics directly ground into models' token probability output \cite{chen2023quantifying, jiang2021can, lin2024contextualized, chen2025first}. These include average log-probability as the mean token log-probabilities, entropy-based measures that quantify prediction uncertainty via sequence or per-token entropy, self-certainty \cite{kang2025scalable} that utilizes the intrinsic probability distribution of LLM outputs to assess the model confidence.

Currently, generative LLMs operate based on an auto-regressive process, formulated as:
\begin{equation}
\begin{aligned}
y_i \sim \text{f}(x,y_{<i}) \ ,
\end{aligned}
\label{eq:ae}
\end{equation}
where $\text{f}()$ is an LLM, $x$ is the prompt, $y_{<i}$ represents the first generated $i-1$ tokens, and $y_i$ denotes the $i^{th}$ token. The output of $\text{f}()$ is the probability distribution over each word in the vocabulary, from which $y_i$ is sampled according to this probability vector. This process can be viewed as a Markov Decision Process (MDP), where the state consists of the prompt $x$ along with the currently generated tokens $y_{<i}$, and the action space corresponds to the vocabulary, allowing each token to be chosen with a probability given by
\begin{equation}
\begin{aligned}
p(v_j|x,y_{<i};\pi_f) = \text{f}(x,y_{<i})[j] \ ,
\end{aligned}
\label{eq:mdp}
\end{equation}
where $v_j$ is the $j^{th}$ element in the vocabulary $v$, and $\pi_f$ represents the strategy. For simplicity, we will neglect the $\pi_f$ term in the following discussion.

\begin{figure}[!t]
\centering
\includegraphics[width=2.5in]{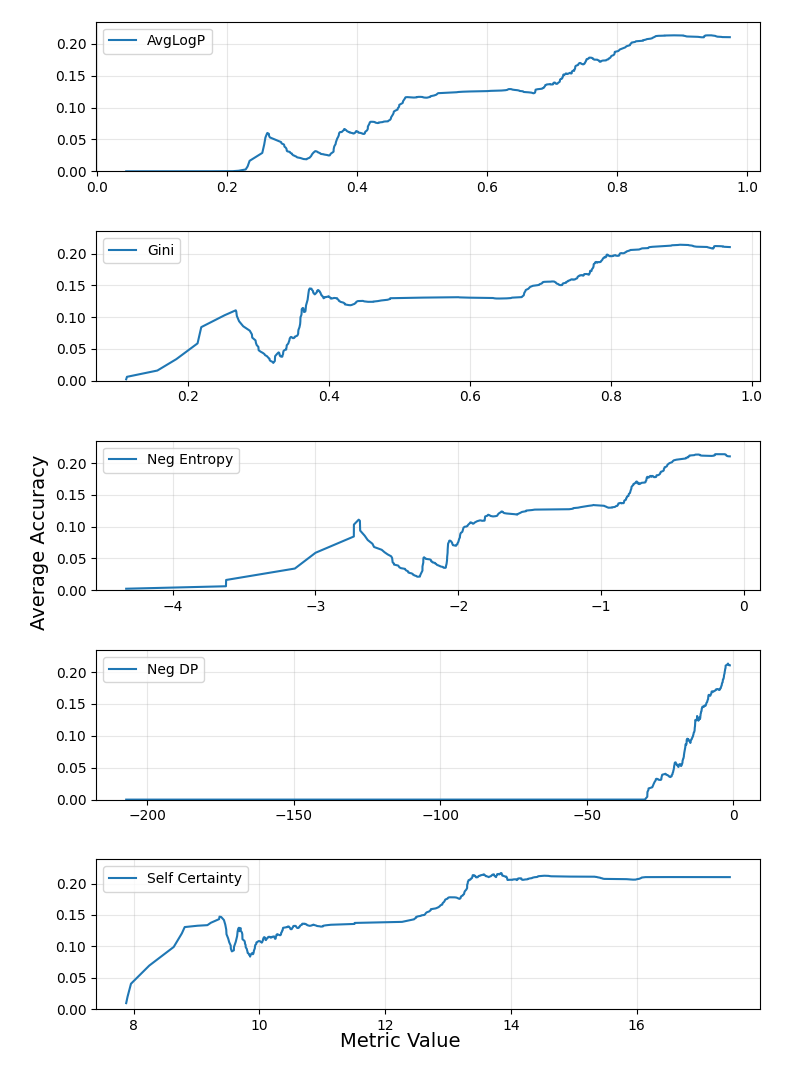}
\caption{CDF of Accuracy for Different Metrics: The lines have been smoothed using a Gaussian filter.}
\label{fig:cdf}
\end{figure}

 Many studies have observed that the probability of each token reflects the LLMs' confidence. Intuitively, if the selected token has a higher chosen probability $p(y_i|x,y_{<i})$, or if the generated probability vector $\text{f}(x,y_{<i})$ exhibits a less uniform distribution, the LLMs demonstrate higher confidence in their generated results. Our study will use the confidence measurement metrics summarized in \cite{kang2025scalable}:

\begin{itemize}
    
    \item \textbf{Average Log-Probability (AvgLogP)}: This metric computes the mean logarithmic likelihood of the generated tokens, reflecting the model's confidence in the entire sequence:
    \begin{equation}
    \text{AvgLogP} = \frac{1}{n}\sum_{i=1}^{n} \log p(y_i|x,y_{<i}) \ ,
    \label{eq:avglogp}
    \end{equation}
    where $n$ represents the answer length. Higher log-probabilities suggest more reliable predictions.

    \item \textbf{Gini Impurity (Gini)}: This metric measures the concentration of the predicted distribution:
    \begin{equation}
    \text{Gini} =  \frac{1}{n}\sum_{i=1}^{n} \sum_{j=1}^{|v|} \left(p(v_j|x,y_{<i})\right)^2 \ ,
    \end{equation}
    where $|v|$ represents the vocabulary size. Higher values indicate more peaked distributions, reflecting greater certainty in the model's predictions.
    
    \item \textbf{Entropy}: This metric computes the uncertainty of the predicted distribution:
    \begin{equation}
    \begin{aligned}
    & \text{Entropy} = \frac{1}{n}\sum_{i=1}^{n} \sum_{j=1}^{|v|}  \\
    & -p(j|x,y_{<i}) \log p(v_j|x,y_{<i}) \ . 
    \end{aligned}
    \label{eq:entropy}
    \end{equation}
    Lower entropy signifies higher confidence, as it indicates a more deterministic output distribution.
    
    \item \textbf{Distributional Perplexity (DP)}: This metric generalizes perplexity to the entire predicted distribution:
    \begin{equation}
    \begin{aligned}
    & \text{DP} = \frac{1}{n}\sum_{i=1}^{n} \exp \\
    & \left(-\sum_{j=1}^{|v|} p(v_j|x,y_{<i}) \log p(v_j|x,y_{<i})\right) \ .
    \end{aligned}
    \label{eq:dp}
    \end{equation}
    Lower values of distributional perplexity indicate that the model is more confident about its predictions across the entire distribution, as it reflects a clearer understanding of token relationships.
    
    \item \textbf{Self-Certainty}: This metric assesses the KL-divergence from a uniform distribution:
    \begin{equation}
    \begin{aligned}
    & \text{Self-certainty} = -\frac{1}{n|v|}\sum_{i=1}^{n}\sum_{j=1}^{V} \\
    & \log\left(|v| \cdot p(v_j|x,y_{<i})\right) \ .
    \end{aligned}
    \label{eq:self-certainty}
    \end{equation}
    Higher self-certainty values reflect greater confidence in the predictions, as they indicate a more concentrated distribution of outputs.
\end{itemize}
The Cumulative Distribution Function (CDF) relationship between these metrics and accuracy is provided in Fig. \ref{fig:cdf}, where we utilize the negative of Entropy and DP.

\setkeys{Gin}{draft=false}
\begin{figure*}[!t]
\centering
\includegraphics[width=\textwidth]{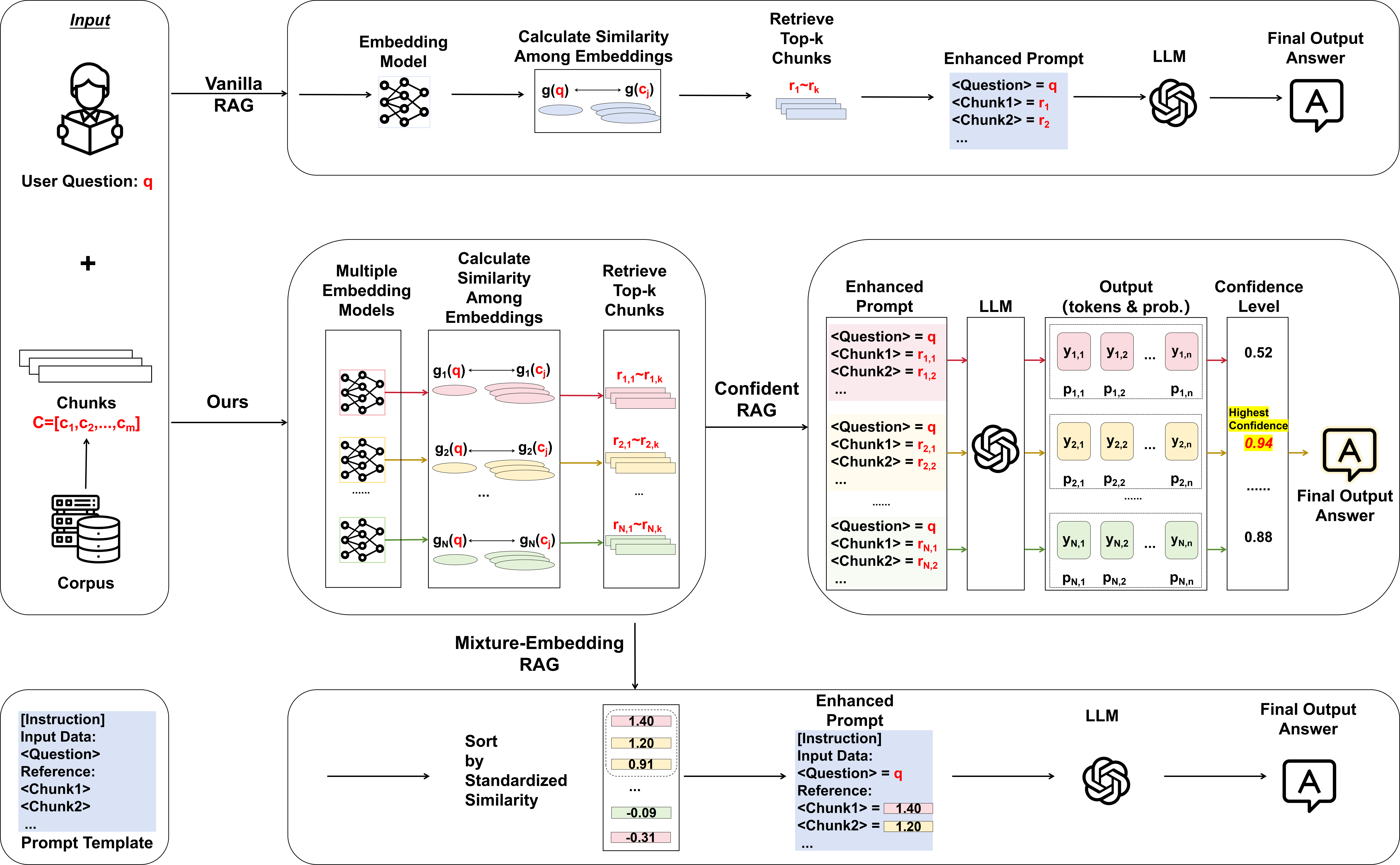}
\caption{Workflow: This figure illustrates the workflows of vanilla RAG, as well as our proposed Mixture-Embedding RAG and Confident RAG.}
\label{fig:main}
\end{figure*}

However, the application of these confidence metrics to mitigate retrieval-based uncertainty remains largely unexplored, particularly in specialized domains like mathematics. Here, a single error can invalidate an entire solution, making reliable confidence scoring essential. Our approach uses confidence as a criterion to select the best output from multiple embedding-grounded generations, directly addressing a key weakness of standard RAG in mathematical contexts.

\subsection{Research Gap and Research Questions}
Previous studies reveal a clear and unaddressed opportunity. While both RAG and confidence estimation are well-studied independently, no prior work has systematically investigated the combination of multiple embedding models with confidence-based answer selection to create a robust, plug-and-play framework for mathematics. To address this gap, two research questions are proposed to guide our study:

Research question 1: Does vanilla RAG provide significant accuracy improvements over vanilla LLMs for mathematical problem-solving?

Research question 2: To what extent do the retrieval-fusion method (Mixture-Embedding RAG) and confidence-selection method (Confident RAG) enhance accuracy rates compared to standard educational AI approaches (vanilla LLM and vanilla RAG) in mathematical problem-solving?

For the Confident RAG method, we have two sub-questions:

Research question 2.1: Which confidence metrics are the most effective for selecting the best final answer in the Confident RAG framework?

Research question 2.2: What is the optimal number of embedding models for the Confident RAG framework?

\section{Methodology}
This study addresses the challenge of unreliable AI in mathematics education by enhancing the Retrieval-Augmented Generation (RAG) framework. Based on the understanding that different embedding models may retrieve different information across the diverse spectrum of mathematical topics encountered by students, we propose two comprehensive RAG methods that combine the strengths of multiple embedding models. Both methods aim to provide students with more accurate and reliable answers by ensuring the AI is grounded in the most relevant mathematical knowledge.

\subsection{Workflow Overview}
In this work, given a student's mathematical question $q$, we utilize a single LLM $\text{f}()$ for answer generation, while $N$ embedding models $\{\text{g}_1(), \text{g}_2(), \ldots, \text{g}_N()\}$ assist in retrieving question-related information from a corpus involving mathematics textbooks and question-answer pairs, which has been divided into chunks $C = [c_{1}, c_{2}, \ldots, c_{m}]$. This setup mirrors a scenario where a student seeks help from a tutor who consults multiple specialized reference books. Inspired by the stacking and bagging paradigms of ensemble learning, we propose two distinct methods: (1) Mixture-Embedding RAG, which first combines the retrieved results from multiple embedding models and incorporates them into the prompt for the LLM, and (2) Confident RAG, which employs vanilla RAG with different embedding models for initial answer generation and subsequently selects the answer with the highest confidence as the final response.

As shown in Fig. \ref{fig:main}, the two proposed methods share the same preliminary process: first, they utilize different embedding models to identify the $k$ chunks with the highest similarity, similar to the approach taken by the vanilla RAG described in Section \ref{sec:RAG}. We define the retrieved chunks as $ R = \{ r_{i,j} \mid i = 1,2,\ldots,N; j = 1,2,\ldots,k \} $, where $r_{i,j}$ is defined as $c_{\mathcal{K}_j^i}$ and $\mathcal{K}^i$ is the index set of the selected $k$ chunks by embedding model $\text{g}_i$. Additionally, we define the similarity between the question $q$ and chunk $r_{i,j}$, calculated by embedding model $\text{g}_i$, as $w_{i,j}$. The differences between the Mixture-Embedding RAG and Confident RAG are detailed as follows.

\subsection{Mixture-Embedding RAG}
In this method, we aim to directly find the chunks with the highest similarity to the question among the candidates selected by all embedding models. The goal is to mimic an expert tutor who synthesizes information from multiple sources to explain a concept. Intuitively, we can select chunks using the similarity score $w_{i,j}$. However, this approach can lead to two problems. First, the candidates from different embedding models may have repetitions. Therefore, we select chunks without repetition to avoid any performance loss. Second, due to the different structures of embedding models, they may have varying ranges of similarity, so that simply comparing similarity across different models may introduce potential bias. To address this, we propose to standardize the similarity score using Z-scores:
\begin{equation}
\hat{w}_{i,j} = \frac{w_{i,j} - \mu_{i}}{\sigma_{i}} \ ,
\label{eq:std}
\end{equation}
where $\mu_i$ and $\sigma_{i}$ represent the mean and standard deviation of all similarity scores from embedding model $\text{g}_i$.

\subsection{Confident RAG}
Inspired by \cite{kang2025scalable}, we first let the LLM $\text{f}$ generate $N$ separate answers, using the retrieved chunks from the $N$ embedding models separately. Then, we evaluate the confidence of the $N$ answers using the metrics mentioned in Section \ref{sec:confidence}. Finally, we select the answer with the highest confidence as our final answer for each question. This method is particularly valuable in educational contexts where a single incorrect solution can significantly impact student learning. By selecting the answer the LLM is most certain about, we provide a safer, more reliable mathematical assistant for students.

\section{Experiment}
\subsection{Experiment Setup}
This section provides an overview of the experiment setup, including the model configuration, dataset, corpus, embedding models, and large language models.

\begin{itemize}
    \item \textbf{Model configuration}: Utilizing six Nvidia A2 GPUs, it took approximately fourteen days to obtian the initial results from our experiment. These included vanilla LLM, vanilla RAG with different embedding models, mixture-embedding RAG with various combinations of embedding models, and the combined results of confident RAG method.
    
    \item \textbf{Dataset}: Gsm8k \cite{cobbe2021gsm8k} is a dataset of approximately 8,500 high-quality, linguistically diverse grade school math word problems designed to support question answering on basic mathematical tasks. These problems typically require 2 to 8 steps of reasoning, mainly involving elementary arithmetic operations, and are solvable by middle school students. The solutions are provided in natural language, emphasizing sequential reasoning rather than complex concepts or variables. The train dataset of Gsm8k was used as the retrieved courpus for RAG, while the first 500 items from the test dataset were used as user questions.
    
    \item \textbf{Corpus}: Two types of corpora were selected for retrieval: (1)  Mathematics textbooks from OpenStax\footnote{\url{https://openstax.org/subjects/math}} covering domains such as calculus, algebra and trigonometry. The content was segmented into sub-section. (2) Math QA items from the Gsm8k train dataset. One textbook sub-section and three QA items will be retrieved according to the similarity based on their similarity to the user-provided math question each time.

    \item \textbf{Embedding models}: We chose four embedding models for encoding, including: all-MiniLM-L6-v2 \cite{all-mini}, ModernBERT-large \cite{modernbert}, MathBERT \cite{peng2021mathbert} and stsb-roberta-large \cite{reimers-2019-sentence-bert} models.

    \item \textbf{LLMs}: To validate our proposed method and ensure repeatability, we selected three LLMs, following \cite{shao2025spuriousrewardsrethinkingtraining}: Qwen2.5-Math-7B \cite{yang2024qwen2}, Llama-3.1-8B \cite{grattafiori2024llama}
    and OLMo-2-1124-7B \cite{olmo20242}. 
\end{itemize}

\subsection{Experiment Result}

\subsubsection{Does vanilla RAG provide significant accuracy improvements over vanilla LLMs for mathematical problem-solving?}

Table \ref{tab:model_performance} presents a comparison of three different LLMs with and without RAG using different embedding models. All models achieve significant improvements in accuracy after applying RAG method, with an average improvement of 5\%. The accuracy of Qwen2.5-Math-7B increases from 75.2\% to 80.5\%, Llama-3.1-8B from 16.6\% to 21.3\%, and OLMo-2-1124-7B from 21.0\% to 26.0\%. This confirms that grounding LLMs in external knowledge is a viable strategy for improving their reliability in mathematical contexts.

However, the performance of vanilla RAG was unstable when using different embedding models. This limitation can be vividly illustrated in Table \ref{tab:model_comparison} for a 400-meter hurdle word problem. The vanilla LLM, without any retrieved context, produced an incorrect answer of 34.2. When augmented with RAG, the results were different: using embedding model 1 or 4 led to the same incorrect answer (34.2), while models 2 and 3 successfully produce the right answer (36).

\begin{table*}[h]
\centering
\begin{threeparttable}

\caption{Model Performance Comparison Across Different Embedding Methods}\label{tab:model_performance}%
\begin{tabular}{lc|cccccc}
\toprule
\multirow{2}{*}{\textbf{LLM Model}} & \multirow{2}{*}{\textbf{Vanilla LLM}} & \multicolumn{6}{c}{\textbf{Vanilla RAG}} \\
 &  & \textbf{Emb1} & \textbf{Emb2} & \textbf{Emb3} & \textbf{Emb4} & \textbf{Avg} & \textbf{Improvement} \\
\midrule
\textbf{Qwen2.5-Math-7B} \cite{yang2024qwen2} & 75.2\% & 81.0\% & 83.0\% & 77.6\% & 80.2\% & 80.5\% & 5.3\%$\uparrow$ \\
\textbf{Llama-3.1-8B} \cite{grattafiori2024llama} & 16.6\% & 21.8\% & 19.8\% & 19.4\% & 24.0\% & 21.3\% & 4.7\%$\uparrow$ \\
\textbf{OLMo-2-1124-7B} \cite{olmo20242} & 21.0\% & 26.0\% & 28.4\% & 25.0\% & 24.6\% & 26.0\% & 5.0\%$\uparrow$ \\
\midrule
\textbf{Average} & 37.6\% &	42.9\%	& 43.7\% &	40.7\% &	42.9\%	& 42.6\% &	5.0\%$\uparrow$\\
\bottomrule

\end{tabular}

\begin{tablenotes}
\small
\item The four embedding models (1–4) correspond to all-MiniLM-L6-v2 \cite{all-mini}, ModernBERT-large \cite{modernbert}, MathBERT \cite{peng2021mathbert}, and stsb-roberta-large \cite{reimers-2019-sentence-bert}, respectively, which will remain consistent in the following tables. The last two columns represent the average accuracy of RAG using the four embedding models and the improvement compared to the vanilla LLM. These details will remain the same in the subsequent tables.
\end{tablenotes}
\end{threeparttable}
\end{table*}

\begin{table*}[h]
\centering
\begin{threeparttable}
\caption{Sample generation results for an item}
\label{tab:model_comparison}
\begin{tabular}{cccccc}
\toprule
\multirow{2}{*}{\textbf{Question}} & \textbf{Correct} & \multirow{2}{*}{\textbf{Model}} & \multirow{2}{*}{\textbf{Answer}} & \textbf{Confidence} & \multirow{2}{*}{\textbf{Accuracy}} \\
 & \textbf{Answer} & & & \textbf{(self-certainty)} & \\
\midrule
\multirow{22}[0]{*}{\begin{tabular}{@{}l@{}} 
Lee used to be able\\ to run the 400-meter\\ hurdles two seconds\\ faster than Gerald\\ would run the 400-\\meter hurdles. But\\ Gerald changed his\\ diet, which improved\\ his speed by 10\%.\\ If Lee runs the 400-\\meter hurdles in 38\\ seconds, how fast\\ can Gerald, with his\\ improved diet, run\\ the 400-meter hurdles,\\in seconds?
\end{tabular}} & \multirow{22}[0]{*}{36} & \textbf{Vanilla LLM} & 34.2 & / & 0 \\
\cmidrule{3-6}
 &  & \textbf{Vanilla RAG} & & & \\
 &  & Emb1 & 34.2 & 9.1794 & 0 \\
 &  & Emb2 & 36 & 16.7625 & 1 \\
 &  & Emb3 & 36 & 15.9693 & 1 \\
 &  & Emb4 & 34.2 & 9.3341 & 0 \\
  \cmidrule{3-6}
 &  & \textbf{Mixture-embedding RAG} & & & \\
 &  & 2 Embs & 36 & / & 1 \\
 &  & 3 Embs & 34.2 & / & 0 \\
 &  & 4 Embs & 34.2 & / & 0 \\
 \cmidrule{3-6}
 &  & \textbf{Confident RAG} & & & \\
 &  & 1,2 & 36 & 16.7625 & 1 \\
 &  & 1,3 & 36 & 15.9693 & 1 \\
 &  & 1,4 & 34.2 & 9.3341 & 0 \\
 &  & 2,3 & 36 & 16.7625 & 1 \\
 &  & 2,4 & 36 & 16.7625 & 1 \\
 &  & 3,4 & 36 & 15.9693 & 1 \\
 &  & 1,2,3 & 36 & 16.7625 & 1 \\
 &  & 1,2,4 & 36 & 16.7625 & 1 \\
 &  & 1,3,4 & 36 & 15.9693 & 1 \\
 &  & 2,3,4 & 36 & 16.7625 & 1 \\
 &  & 1,2,3,4 & 36 & 16.7625 & 1 \\
\bottomrule
\end{tabular}

\begin{tablenotes}
\small
\item The LLM used in this example is Qwen2.5-Math-7B. The four embedding models (1-4) are the same as the previous table. The Mixture-embedding RAG results illustrate the performance when using retrieved information from 2 to 4 different embedding models randomly (denoted as 2 Embs to 4 Embs). The Confident RAG results illustrate the performance when using different combinations of multiple embedding models and using self-certainty as the confidence metric. Accuracy: 0=incorrect; 1=correct.
\end{tablenotes}
\end{threeparttable}
\end{table*}

\subsubsection{To what extent do Mixture-Embedding RAG enhance accuracy than vanilla LLM and RAG in mathematical problem-solving?}

The Mixture-Embedding RAG method aimed to solve this instability by fusing retrieved documents. However, as shown in Table \ref{tab:mix_embedding RAG}, the average accuracies of the three LLMs did not perform well compared to vanilla RAG. Llama-3.1-8B and OLMo-2-1124-7B demonstrated similar accuracy as vanilla RAG, with the differences ranging from -0.2\% to 0.5\%. Surprisingly, Qwen2.5-Math-7B showed a decrease of 5.5\% compared to vanilla RAG.

The example in Table \ref{tab:model_comparison} reveals the potential pedagogical danger of this approach. While using 2 embedding models (2 Embs) yielded the correct answer (36), fusing 3 or 4 embedding models (3 Embs, 4 Embs) resulted in incorrect answer (34.2). This suggests that the fusion process can sometimes amplify noisy or incorrect information, leading the LLM to a wrong conclusion and therefore resulting in a critical failure for educational tools designed to build correct conceptual understanding.

\begin{table*}[h]
\centering
\begin{threeparttable}

\caption{Performance of Mixture-Embedding RAG}\label{tab:mix_embedding RAG}
\begin{tabular}{lcccccc}
\toprule
\multirow{2}{*}{\textbf{LLM Model}} & \multicolumn{6}{c}{\textbf{Mix-Embedding RAG}} \\
 &    \textbf{2 Embs} & \textbf{3 Embs} & \textbf{4 Embs}	& \textbf{Avg} & \textbf{v.s. Vanilla LLM} & \textbf{v.s. Vanilla RAG}
\\
\midrule
\textbf{Qwen2.5-Math-7B} \cite{yang2024qwen2}  & 74.20\% &	76.00\% &	74.80\% &	75.0\% &	-0.2\%$\downarrow$ &	-5.5\%$\downarrow$
\\
\textbf{Llama-3.1-8B} \cite{grattafiori2024llama}  & 20.90\% &	19.60\% &	22.80\% &	21.1\% &	4.5\%$\uparrow$ &	-0.2\%$\downarrow$
\\
\textbf{OLMo-2-1124-7B} \cite{olmo20242}  & 26.20\%	& 26.80\% &	26.40\% &	26.5\% &	5.5\%$\uparrow$ &	0.5\%$\uparrow$
\\
\midrule
\textbf{Average}  & 40.4\%	& 40.8\% &	41.3\% &	40.9\% &	3.3\%$\uparrow$ &	-1.7\%$\downarrow$
\\
\bottomrule
\end{tabular}

\begin{tablenotes}
\small
\item This table illustrates the performance when using retrieved information from 2 to 4 different embedding models randomly (denoted as 2 Embs to 4 Embs), comparing the results with those of the vanilla LLM and vanilla RAG.
\end{tablenotes}
\end{threeparttable}

\end{table*}

\subsubsection{To what extent do Confident RAG enhance accuracy than vanilla LLM and RAG in mathematical problem-solving?}

The Confident RAG method demonstrated superior and consistent performance, addressing the core instability of previous approaches. After using different embedding models for RAG in multiple rounds, the results of Confident RAG method are shown in Table \ref{tab:combine_performance}. Four main findings were obtained:

(1) The accuracy of Confident RAG method surpasses that of both the vanilla RAG (with an average improvement ranging from 3.2\% to 4.9\%) and vanilla LLM (with an average improvement ranging from 8.1\% to 9.9\%). 

(2) For each LLM, after applying the Confident RAG method, the accuracy improved by nearly 10\% when using the best confidence metric compared to the vanilla LLM. 

(3) Among all confidence metrics, self certainty and distributional perplexity demonstrated the best performance, with average improvements of 9.9\% and 9.7\%, respectively, over the vanilla LLM. These two metrics also performed well across different LLMs. For instance, there was an increase of 9.1\% for Qwen2.5-Math-7B using self certainty as the confidence metric, an increase of 10.4\% for Llama-3.1-8B with both metrics, and an increase of 12.3\% for OLMo-2-1124-7B using distributional perplexity. 

The practical mechanism behind this success is clearly demonstrated in our running example (Table \ref{tab:model_comparison}). For the hurdle problem, Confident RAG generated multiple candidate answers. Crucially, the self-certainty metric correctly assigned a high score (16.76) to the correct answer (36) and significantly lower scores (9.3) to the incorrect one (34.2). In every combination of embedding models, we selected the answer with the high-confidence, which can largely minimizing the risk of propagating pedagogical errors and present the most reliable solution to the students. 

(4) Regarding the optimal number of embedding models (N) for multi-rounds, no evidence suggests that a larger n yields better accuracy. In our experiments, the accuracy when N=3 was always larger than when N=2. Meanwhile, the accuracy for N=3 and N=4 was similar, with a maximum difference of 1\% across the three LLMs. Considering factors such as time cost, GPU capacity, and other constrains, this finding suggests that researchers should seek a suitable trade-off based on their own condition when choosing the optimal N.

\begin{table*}[htbp]
    \centering 
    \begin{threeparttable}
    \caption{Accuracy Comparison Across Multi-RAG with Different Embedding Models}\label{tab:combine_performance}%
    \begin{tabular}{ll|cccccc}
        \toprule
        \textbf{LLM} & \textbf{Emb. Model} 
        & \textbf{AvgLogP} & \textbf{Self-certainty} & \textbf{Gini} & \textbf{Entropy} & \textbf{DP} \\ 
        \midrule
        \multirow{16}{*}{\textbf{Qwen2.5-Math-7B} \cite{yang2024qwen2}}  & 1,2 & \underline{82.0\%} & \underline{\textbf{85.0\%}} & \underline{81.8\%} & \underline{82.8\%} & \underline{83.6\%} \\ 
         & 1,3 & 79.4\% &\textbf{83.4\%} & 79.2\% & 79.6\% & 81.0\% \\ 
         & 1,4 & 81.4\% & \textbf{84.6\%} & 81.6\% & 81.8\% & 82.2\% \\ 
         & 2,3 & 81.8\% & \textbf{84.4\%} & 81.4\% & 81.8\% & 82.4\% \\ 
         & 2,4 & 81.4\% & \textbf{83.6\%} & 81.0\% & 81.4\% & 82.2\% \\ 
         & 3,4 & 79.2\% &\textbf{82.2\%} & 79.8\% & 79.8\% & 80.6\% \\ 
         & Avg (n=2) &	80.9\% &	\textbf{83.9\%} &	80.8\% &	81.2\% &	82.0\% \\
         \cmidrule{2-7}
         & 1,2,3 & 79.4\% & \underline{\textbf{85.0\%}} & 79.0\% & 79.8\% & 81.6\% \\ 
         & 1,2,4 & 79.6\% & \underline{\textbf{85.0\%}} & 79.6\% & 80.6\% & 82.0\% \\ 
         & 1,3,4 & 79.0\% & \textbf{84.8\%} & 79.6\% & 80.0\% & 80.8\% \\ 
         & 2,3,4 & 79.2\% & \textbf{84.4\%} & 79.0\% & 79.8\% & 81.0\% \\ 
         & Avg (n=3)	& 79.3\% &	\textbf{84.8\%} &	79.3\% &	80.1\% &	81.4\% \\
         \cmidrule{2-7}
         & 1,2,3,4 & 78.2\% & \textbf{84.8\%} & 78.4\% & 79.2\% & 80.6\% \\ 
         \cmidrule{2-7}
         & Avg (n=2,3,4) & 80.1\% & \textbf{84.3\%} & 80.0\% & 80.6\% & 81.6\% \\ 
         & v.s. Vanilla RAG & 0.4\%$\downarrow$ &	\textbf{3.8\%}$\uparrow$ &	0.4\%$\downarrow$ & 0.1\%$\uparrow$	& 1.2\%$\uparrow$ \\
         & v.s. Vanilla LLM & 4.9\%$\uparrow$ &	\textbf{9.1\%}$\uparrow$ &	4.8\%$\uparrow$ &	5.4\%$\uparrow$ &	6.4\%$\uparrow$ \\
        \hline \hline
        \multirow{16}{*}{\textbf{Llama-3.1-8B} \cite{grattafiori2024llama}} & 1,2 & \textbf{27.2\%} & \textbf{27.2\%} & 27.0\% & 27.0\% & \textbf{27.2\%} \\ 
         & 1,3 & 26.8\% & 26.8\% & 26.6\% & 27.0\% & \textbf{27.2\%} \\ 
         & 1,4 & \textbf{26.8\%} & 26.4\% & 26.4\% & 26.4\% & 26.4\% \\ 
         & 2,3 & 23.2\% & \textbf{24.0\%} & \textbf{24.0\%} & \textbf{24.0\%} & 23.6\% \\ 
         & 2,4 & 26.6\% & \textbf{26.8\%} & 26.4\% & 26.0\% & 26.6\% \\ 
         & 3,4 & 25.8\% & 26.4\% & \textbf{26.6\%} & \textbf{26.6\%} & 26.4\% \\ 
         & Avg (n=2) &	26.1\% &	\textbf{26.3\%} &	26.2\% &	26.2\% &	26.2\% \\
         \cmidrule{2-7}
         & 1,2,3 & 28.6\% & \underline{29.2\%} & \underline{28.8\%} & \underline{\textbf{29.4\%}} & \underline{\textbf{29.4\%}} \\ 
         & 1,2,4 & \underline{\textbf{28.8\%}} & 28.6\% & 28.0\% & 28.0\% & 28.2\% \\ 
         & 1,3,4 & 27.4\% & 27.4\% & 27.2\% & \textbf{27.6\%} & \textbf{27.6\%} \\ 
         & 2,3,4 & 25.8\% & \textbf{26.4\%} & 26.0\% & 26.0\% & \textbf{26.4\%} \\ 
         & Avg (n=3)	& 27.7\% &	\textbf{27.9\%} &	27.5\% &	27.8\% &	\textbf{27.9\%} \\
         \cmidrule{2-7}
         & 1,2,3,4 & \textbf{27.8\%} & 27.6\% & 27.0\% & 27.4\% & 27.6\% \\ 
         \cmidrule{2-7}
         & Avg (n=2,3,4) & 26.8\% & \textbf{27.0\%} & 26.7\% & 26.9\% & \textbf{27.0\%} \\
         & v.s. Vanilla RAG & 5.6\%$\uparrow$ &	\textbf{5.7\%}$\uparrow$ &	5.5\%$\uparrow$ &	5.6\%$\uparrow$ &	\textbf{5.7\%}$\uparrow$ \\
         & v.s. Vanilla LLM & 10.2\%$\uparrow$ &	\textbf{10.4\%}$\uparrow$ &	10.1\%$\uparrow$ &	10.3\%$\uparrow$ &	\textbf{10.4\%}$\uparrow$ \\
        \hline \hline
        \multirow{16}{*}{\textbf{OLMo-2-1124-7B} \cite{olmo20242}} & 1,2 & 31.0\% & 31.2\% & 30.4\% & 31.2\% & \textbf{32.2\%} \\ 
         & 1,3 & 29.8\% & 29.6\% & 29.8\% & \textbf{30.0\%} & \textbf{30.0\%} \\ 
         & 1,4 & 29.4\% & 29.2\% & 28.6\% & 29.6\% & \textbf{31.0\%} \\ 
         & 2,3 & 31.6\% & 30.8\% & 30.2\% & 31.4\% & \textbf{32.8\%} \\ 
         & 2,4 & 32.6\% & 32.0\% & 31.0\% & \textbf{33.0\%} & 33.8\% \\ 
         & 3,4 & 29.6\% & 30.2\% & 29.6\% & 30.6\% & \textbf{31.4\%} \\ 
         & Avg (n=2) &	30.7\% &	30.5\% &	29.9\%	& 31.0\% &	\textbf{31.9\%} \\
         \cmidrule{2-7}
         & 1,2,3 & 32.6\% & 32.0\% & 31.4\% & 32.2\% & \textbf{33.2\%} \\ 
         & 1,2,4 & \underline{32.8\%} & \underline{32.8\%} & \underline{31.6\%} & 33.2\% & \textbf{34.8\%} \\ 
         & 1,3,4 & 31.2\% & 31.2\% & 30.6\% & 32.0\% & \textbf{34.6\%} \\ 
         & 2,3,4 & 32.6\% & 32.2\% & 30.8\% & \underline{33.6\%} & \underline{\textbf{36.6\%}} \\ 
         & Avg (n=3)	& 32.3\% &	32.1\% &	31.1\% &	32.8\% &	\textbf{34.8\%} \\
         \cmidrule{2-7}
         & 1,2,3,4 & \underline{32.8\%} & 32.4\% & 31.0\% & 33.2\% & \textbf{35.8\%} \\ 
         \cmidrule{2-7}
         & Avg (n=2,3,4) & 31.5\% & 31.2\% & 30.5\% & 31.8\% & \textbf{33.3\%} \\
         & v.s. Vanilla RAG & 5.5\%$\uparrow$ &	5.2\%$\uparrow$ &	4.5\%$\uparrow$ &	5.8\%$\uparrow$ &	\textbf{7.3\%}$\uparrow$  \\
         & v.s. Vanilla LLM & 10.5\%$\uparrow$ &	10.2\%$\uparrow$ &	9.5\%$\uparrow$ &	10.8\%$\uparrow$ &	\textbf{12.3\%}$\uparrow$ \\
        \hline \hline
         \multirow{6}{*}{\textbf{Average}} & Avg (n=2) &	45.9\% &	\textbf{46.9\%} &	45.6\% &	46.1\% &	46.7\% \\
         & Avg (n=3) &	46.4\% &	\textbf{48.3\%}	 & 46.0\% &	46.9\% &	48.0\% \\
         & Avg (n=4) &	46.3\%	& \textbf{48.3\%} &	45.5\% &	46.6\% &	48.0\% \\
         & Avg (n=2,3,4) & 46.1\% & \textbf{47.5\%} & 45.7\% & 46.4\% & 47.3\% \\ 
         & v.s. Vanilla RAG 	& 3.5\%$\uparrow$ &	\textbf{4.9\%}$\uparrow$ &	3.2\%$\uparrow$ &	3.9\%$\uparrow$ &	4.7\%$\uparrow$ \\
         & v.s. Vanilla LLM	& 8.5\%$\uparrow$ &	\textbf{9.9\%}$\uparrow$ &	8.1\%$\uparrow$ &	8.8\%$\uparrow$ &	9.7\%$\uparrow$ \\
        \bottomrule
        
    \end{tabular}
\begin{tablenotes}
\small
\item Avg($n$) denotes the average accuracy across different combinations of $n$ embedding models. Each column uses \underline{underline} to indicate the best embedding combination within each LLM. Each row uses \textbf{bold} to signify the best metric for confidence evaluation. For performance comparison with Vanilla RAG, we use the average accuracy of all single embedding models as the baseline.
\end{tablenotes}
\end{threeparttable}
    
\end{table*}

\section{Discussion}
\subsection{Mixture-embedding RAG Showed Limited Improvements from Vanilla RAG}

Our results indicate that Mixture-Embedding RAG did not provide the anticipated improvement. For general LLMs (e.g., Llama-3.1-8B and OLMo-2-1124-7B) without math fine-tuning, their internal math knowledge is limited, leading to lower accuracy in direct answer generation \cite{lewkowycz2022solving}. In these cases, even noisy references retrieved by RAG are more reliable than the LLMs' own outputs, as RAG at least provides partially correct information. However, while the mixture-embedding RAG method may optimize the retrieval ranking process and improve the quality of the references, the general LLMs' capabilities prevent them from fully leveraging higher-quality references, resulting in performance similar to vanilla RAG. Additionally, if different embedding models return highly diverse references, directly combining the top-ranked documents may cause information overload or contextual confusion, negating the potential benefits of mixture-embedding method. Therefore, the performance of general LLMs matches that of vanilla RAG rather than surpassing it.

On the other hand, for the LLMs that have been fine-tuned based on math corpora (e.g., Qwen2.5-Math-7B), vanilla RAG may result in smaller improvements, and lower-quality references can lead to poorer answer performance \cite{lewis2020retrieval}. For these types of LLMs, the mixture-embedding method may introduce additional noise in mathematical contexts, resulting in lower accuracy compared to vanilla RAG. The decmidrule in accuracy may be caused by several factors: (1) Mathematical symbols and formulas may vary drastically across embedding models, making similarity calculations unstable. (2) Different models may encode math terms differently, causing the top-ranked reference to be suboptimal. (3) An embedding model might incorrectly rank an irrelevant math material highly, while better references from other models are ignored. If the LLM generates hallucinated answers based on incorrect references, its performance can degrade below that of the vanilla LLM. (4) Information overload or contextual confusion may also occur, similar to what happens with general LLMs.

\subsection{Confident RAG Demonstrated Consistent Improvements through Confidence Selection}

As shown in Figure \ref{fig:cdf}, there exists a positive correlation between confidence and accuracy, which also aligns with the finding from \cite{chen2025first}. Therefore, the Confident RAG method improves overall accuracy by integrating multiple embedding models to generate answers and selecting the highest-confidence results using the most effective metric. This process effectively filters out low-confidence incorrect answers. Meanwhile, the combined effect of RAG and confidence filtering enhances the robustness, leading to significant improvements compared to vanilla LLMs. When employing the optimal confidence metric, all LLMs achieved an accuracy increase of nearly 10\%, demonstrating the method's universality. In the experiments, when the number of embedding models N > 3, the accuracy improvement became limited, likely due to redundant or noisy retrievals introduced by additional models. At N=3, the method achieved an optimal balance between diversity and computational efficiency. Increasing the number of models further yields only marginal benefits.

Self-Certainty and DP outperform other metrics since they directly measure the concentration and divergence of the probability distribution. Specifically, Self-Certainty measures how far the predicted distribution deviates from uniform. By scaling the probabilities by $|v|$ and taking the negative logarithm, it heavily penalizes uniform-like distributions, favoring sharp peaks. This makes it highly discriminative for high-confidence answers. Additionally, DP is an exponential version of entropy. The exponentiation amplifies differences in entropy, making it more sensitive to the sharpness of the distribution. Low DP values indicate tightly clustered high-probability tokens, which strongly correlate with correct answers. In contrast, other metrics are less sensitive because they either average out uncertainties (AvgLogP) or lack normalization across vocabularies (Gini). While entropy can be useful, it is linear and less discriminative compared to DP’s exponential scaling. Therefore, Self-Certainty and DP are more sensitive to subtle variations in model confidence.

\subsection{Practical Implications}
The findings of this study lead to the following three practical implications. 

First, our work on Confident RAG establishes a critical foundation for the future deployment of fully autonomous Agentic RAG systems in educational settings. Leverages multiple embedding models and a confidence-based selection mechanism, our results indicate that it can achieve an accuracy improvement of approximately 10\% compared to the vanilla LLM when using a self-certainty or DP confidence metric. 

For autonomous planning \cite{huang2024understanding}, the agentic system could dynamically determine the optimal number of embedding models to employ based on the confidence scores, and select specific embedding models from a broader bank according to their historical performance. For iterative refinement \cite{huang2023towards}, the system could automatically initiate additional rounds of retrieval using different embedding models if the initial results do not meet a predefined confidence standard, ensuring high-quality output. For long-term memory \cite{zhang2025survey}, the system could keep a learning record, remembering which embedding models work best for different kinds of math problems, allowing it to become smarter and more efficient over time.

Second, this research gives educators and schools a simple way to judge AI teaching tools. They should prioritize tools that demonstrate a methodology for cross-verification and confidence estimation. Our Confident RAG method is an example of this safer and more reliable approach, which is crucial for students' learning.

Third, the Confident RAG method supports students' independent learning and saves teachers' time. With a more trustworthy AI, students can do their homework and explore topics on their own with less risk of learning from a wrong answer. This builds their confidence in self-directed learning. At the same time, it reduces the heavy burden on teachers. They will spend less time double-checking every AI-generated answer for mistakes, freeing them up to focus on actual teaching and helping students who need it most.

\section{Limitations and Future Study}
This study has two main limitations. First, the experiment was only designed for mathematics domain using the GSM8K benchmark and a corpus of mathematical textbooks. This design prevents generalization of our findings to other subjects in educational context. Future study should validate the Confident RAG framework across diverse academic domains. This will test its robustness and determine whether it is a truly general-purpose solution for educational AI.

Second, this study provides a validated methodological framework (Confident RAG) that researchers and developers can reference to build more reliable AI math assistants. However, we have not yet embedded this method into a complete, user-facing educational system. Therefore, its usability and practical impact in a real learning environment remain unexplored. Future study should build a functional AI math tutoring system that incorporates the Confident RAG framework. This will allow us to validate the method's effectiveness not just on benchmark accuracy, but on its ability to support student learning through interactive use, and to refine its design based on user feedback from actual classrooms.

\section{Conclusion}
In this paper, we proposed two novel RAG-enhanced approaches using multiple embedding models to improve the mathematics question response performance of LLMs: (1) Mixture-Embedding RAG and (2) Confident RAG. Our findings clearly indicate that Confident RAG is a superior and robust solution, with an average improvement of approximately 10\% while utilizing the best confidence metric, compared to vanilla LLM. It selects the highest-confidence response from multiple RAG outputs, each generated using a different embedding model. The consistent success of confidence metrics like self-certainty and DP provides a reliable mechanism for identifying correct answers. Crucially, the number of embedding models (N) can be determined by researchers based on their specific conditions, as no evidence suggesting that a larger N necessarily yields a better performance. In contrast, Mixture-Embedding RAG showed no improvement over vanilla RAG, by sorting and selecting retrieved references from different embedding models based on their standardized similarity before guiding LLM. The findings indicate that Confident RAG can be served as a plug-and-play method for enhancing LLM performance in mathematics problem-solving and building more accurate and reliable Agentic RAG system in mathematics education.


\bibliographystyle{IEEEtran}
\bibliography{bibliography}

\end{document}